\documentclass[sigconf]{acmart}

\usepackage{balance}
\usepackage{booktabs}
\usepackage{multirow}

\def\BibTeX{{\rm B\kern-.05em{\sc i\kern-.025em b}\kern-.08emT\kern-.1667em\lower.7ex\hbox{E}\kern-.125emX}}
    
\copyrightyear{2019}
\acmYear{2019}
\setcopyright{acmcopyright}
\acmConference[SIGIR '19]{Proceedings of the 42nd International ACM SIGIR Conference on Research and Development in Information Retrieval}{July 21--25, 2019}{Paris, France}
\acmBooktitle{Proceedings of the 42nd International ACM SIGIR Conference on Research and Development in Information Retrieval (SIGIR '19), July 21--25, 2019, Paris, France}
\acmPrice{15.00}
\acmDOI{10.1145/3331184.3331351}
\acmISBN{978-1-4503-6172-9/19/07}

\begin{document}

\title{Syntax-Aware Aspect-Level Sentiment Classification with Proximity-Weighted Convolution Network}

\author{Chen Zhang}
\affiliation{
  \institution{Beijing Institute of Technology}
  \city{Beijing}
  \country{China}
}
\affiliation{
  \institution{\& Zhejiang Lab}
  \city{Hangzhou}
  \country{China}
}
\email{gene@bit.edu.cn}
\author{Qiuchi Li}
\affiliation{%
  \institution{University of Padua}
  \city{Padua}
  \country{Italy}
}
\email{qiuchili@dei.unipd.it}
\author{Dawei Song}
\authornote{Corresponding author.}
\affiliation{%
  \institution{Beijing Institute of Technology}
  \city{Beijing}
  \country{China}
}
\email{dwsong@bit.edu.cn}

%
\renewcommand{\shortauthors}{C. Zhang et al.}

%
\begin{abstract}
It has been widely accepted that Long Short-Term Memory (LSTM) network, coupled with attention mechanism and memory module, is useful for aspect-level sentiment classification. However, existing approaches largely rely on the modelling of semantic relatedness of an aspect with its context words, while to some extent ignore their syntactic dependencies within sentences. Consequently, this may lead to an undesirable result that the aspect attends on contextual words that are descriptive of other aspects. In this paper, we propose a proximity-weighted convolution network to offer an aspect-specific syntax-aware representation of contexts. In particular, two ways of determining proximity weight are explored, namely position proximity and dependency proximity. The representation is primarily abstracted by a bidirectional LSTM architecture and further enhanced by a proximity-weighted convolution. Experiments conducted on the SemEval 2014 benchmark demonstrate the effectiveness of our proposed approach compared with a range of state-of-the-art models\footnote{Code is available at \href{https://github.com/GeneZC/PWCN}{https://github.com/GeneZC/PWCN}.}.
\end{abstract}

%
%


%

%
\maketitle

\section{Introduction}

Aspect-level sentiment classification, also called aspect-based sentiment classification, is a fine-grained sentiment classification task aiming at identifying the polarity of a given aspect within a certain context, i.e. a comment or a review. For example, in the following comment about food ``\textit{They use fancy ingredients, but even fancy ingredients don't make for good pizza unless someone knows how to get the crust right.}'', the sentiment polarities for aspects \textit{ingredients}, \textit{pizza} and \textit{crust} are \textit{positive}, \textit{negative} and \textit{neutral} respectively.

Aspect-level sentiment classification has attracted an increasing attention in the fields of Natural Language Processing (NLP) and Information Retrieval (IR), and plays an important role in various applications such as personalized recommendation. Earlier works in this area focused on manually extracting refined features and feeding them into classifiers like Support Vector Machine (SVM)~\cite{Jiang:2011:TTS:2002472.2002492}, which is labor intensive. In order to tackle the problem, automatic feature extraction has been investigated. For example, ~\citet{P14-2009} proposed to adaptively propagate the sentiments of context words to the aspect via their syntactic relationships. ~\citet{Vo:2015:TTS:2832415.2832437} built a syntax-free feature extractor to identify a rich source of relevant features. Despite the effectiveness of these approaches, Tang et al.~\cite{C16-1311} claimed that the modelling of semantic relatedness of an aspect and its context remained a challenge, and proposed to use target-dependent LSTM network to address this challenge. 

As the attention mechanism and memory network have yielded good results in many NLP tasks such as machine translation~\cite{D15-1166,BahdanauCB14}, LSTM combined with attention~\cite{Ma:2017:IAN:3171837.3171854, C18-1096} or memory network~\cite{D16-1021, D17-1047} is deployed to aspect-level sentiment classification to aggregate contextual features for prediction. Being capable of modelling semantic interactions between aspects and their corresponding contexts, these models have improved performance over previous approaches. However, they generally ignore the syntactic relations between the aspect and its context words, which may hinder the effectiveness of aspect-based context representation. For instance, a given aspect may attend on several context words that are descriptively near to the aspect but not correlated to the aspect syntactically. As a concrete example, in ``\textit{Its size is ideal and the weight is acceptable.}'', the aspect term \textit{size} may easily be depicted by \textit{acceptable} based on the semantic relatedness, which is in fact not the case. Syntactic parsing has been used in some previous work ~\cite{P14-2009}, however, the word-level parsing could impede feature extraction across different phrases, as the sentiment polarity of an aspect is usually determined by a key phrase instead of a single word~\cite{Fan:2018:CMN:3209978.3210115}.

In order to address the limitations mentioned above, we propose an aspect-level sentiment classification framework that leverages the syntactic relations between an aspect and its context and aggregates features at the n-gram level, within a LSTM-based architecture. Inspired by the position mechanism~\cite{D16-1021, D17-1047, P18-1087, C18-1096}, the framework utilizes a context word's syntactic proximity to the aspect, a.k.a proximity weight, to determine its importance in the sentence. We then integrate the proximity weights into a convolution network to capture n-gram information, called as \textbf{Proximity-Weighted Convolution Network} (PWCN). Finally, a layer of max-pooling is adopted to select the most significant features for prediction.

Experiments are conducted on SemEval 2014 Task4 datasets. The results show that our model achieves a higher performance than a range of state-of-the-art models, and hence illustrate that syntactical dependencies are more beneficial than semantic relatedness to aspect-level sentiment classification.
\section{The Proposed Model}

\begin{figure}[]
\centering
\includegraphics[width=0.47\textwidth]{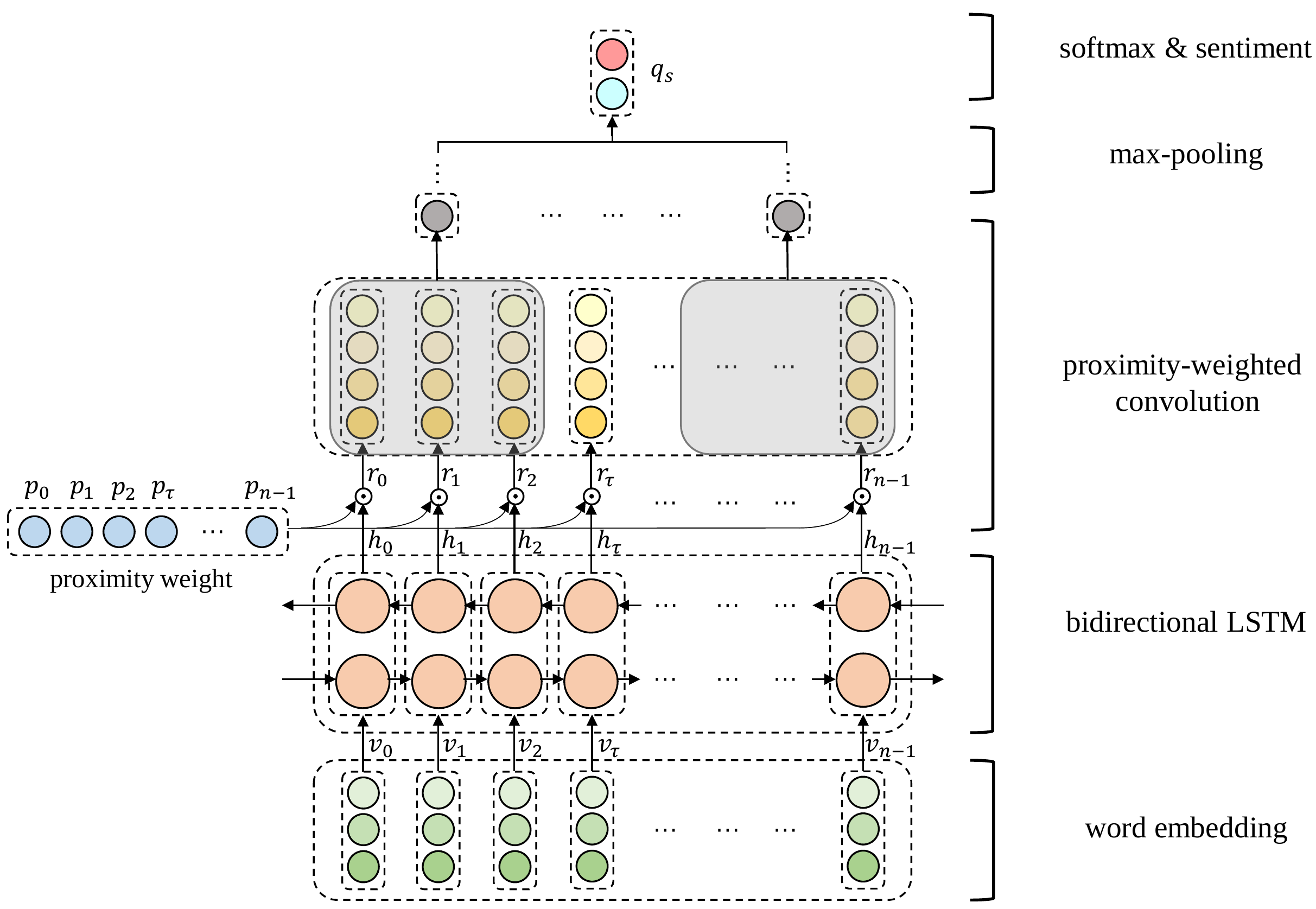}
\caption{Overview of the model architecture.}
\label{fig:fig1}
\end{figure}

An overview of our proposed model is given in Figure \ref{fig:fig1}. In the model, an $n$-word sentence containing a target $m$-word aspect term is formulated as $\mathbf{S}=\{w_0,w_1, \dots,w_{\tau},w_{\tau+1},\dots,w_{\tau+m-1},\dots,$ $w_{n-1}\}$, where $\tau$ denotes the start token of the aspect term. Each word is embedded into a low-dimensional real-valued vector with a matrix $\mathbf{E}\in\mathbb{R}^{|V|\times d_e}$~\cite{bengio2003neural}, where $|V|$ is the size of dictionary while $d_e$ is the dimensionality of a word vector. After word vectors $\mathbf{V}=\{e_0,e_1,\dots,e_{\tau},e_{\tau+1},\dots,e_{\tau+m-1},\dots,e_{n-1}\}$ are obtained through word embedding, a bidirectional LSTM is adopted to produce the hidden state vectors $\mathbf{H}=\{h_0,h_1,\dots,h_{\tau},h_{\tau+1},\dots,h_{\tau+m-1},\dots,$ $h_{n-1}\}$. Particularly, $h_i\in \mathbb{R}^{2d_h}$ is a concatenation of hidden states respectively obtained from the forward LSTM and the backward LSTM, where $d_h$ is the dimensionality of a hidden state vector in an unidirectional LSTM. The hidden state representation is further enhanced by proximity-weighted convolution and then used for prediction of sentiment polarity. 

\subsection{Proximity Weight}

Previous attention-based models mainly focus on how to obtain a context representation based on its component words' semantic correlations with a corresponding aspect~\cite{Ma:2017:IAN:3171837.3171854, C18-1096, D16-1021, D17-1047, P18-1087, Fan:2018:CMN:3209978.3210115}. These models calculate attention weights referring to word vector representation in the latent semantic space, without taking into consideration syntax information. This may limit the effectiveness of these models in term of mis-identify crucial context words for characterizing the aspect. Therefore, we replace this complicated modelling of aspects by incorporating syntactical dependencies to uncover component words' characteristics to the aspect\footnote{We have conducted experiments using proximity weight combined with attention weight, i.e. the combination of semantic relatedness and syntactic proximity, but we get unexpected sub-optimal results, which will be shown in experiments section.}. Such syntactical dependency information in our proposed model is formalized as \textit{proximity weight}, which describes the contextual words' proximity to the aspect. Recall the example related to \textit{weight} of a laptop saying that ``\textit{Its size is ideal and the weight is acceptable.}''. The set of words including \{\textit{ideal}, \textit{acceptable}\} that are closer to the aspect term \textit{weight} in terms of semantics, should have a larger probability describing the weight of a laptop. Further, from the perspective of syntax parsing, \textit{ideal} could be safely excluded from the word set as it is syntactically too far from \textit{weight}. Actually, \textit{acceptable} is the true descriptor of \textit{weight}, indicating a positive sentiment.  

Following this idea, we propose two different methods, namely \textit{position proximity} and \textit{dependency proximity}, to model the syntactical dependency between contextual words and the aspect term respectively. 

\subsubsection{Position Proximity} 

Generally, it is more likely to see that words around an aspect are describing the aspect. Thus, we view such position information as an approximated syntactical proximity measurement. Position proximity weights are computed by the formula below:
\begin{equation}
p_i=\begin{cases}
1-\frac{\tau-i}{n}& 0 \leq i<\tau\\
0& \tau \leq i < \tau+m\\
1-\frac{i-\tau-m+1}{n}& \tau+m \leq i < n
\end{cases}
\end{equation}

\noindent where proximity weight $p_i \in \mathbb{R}$. Intuitively, the weight decreases in proportion to the word's distance to the nearest border of the aspect term.

\subsubsection{Dependency Proximity} 

Apart from the absolute position in the context, we also consider measuring the distances between words in a syntax dependency parsing tree. For example, in a comment ``\textit{the food is awesome - definitely try the striped bass.}'' with \textit{food} as the aspect, we first construct a dependency tree\footnote{With spaCy toolkit: https://spacy.io/.}, then compute for a context word the tree-based distance, i.e. the length of the shortest path in the tree, between the word and \textit{food}. If the aspect otherwise contains more than one word, we take the minimum of the tree-based distances between a context word and all the aspect component words. In the uncommon case where more than one dependency trees are present in a context, we manually set the distance between the aspect term and context words in other trees to a constant, i.e. half of the sentence length\footnote{It's a proper number that could serve as the boundary of possible descriptive contextual words in our experiments.}.

\begin{figure}[]
\centering
\includegraphics[width=0.47\textwidth]{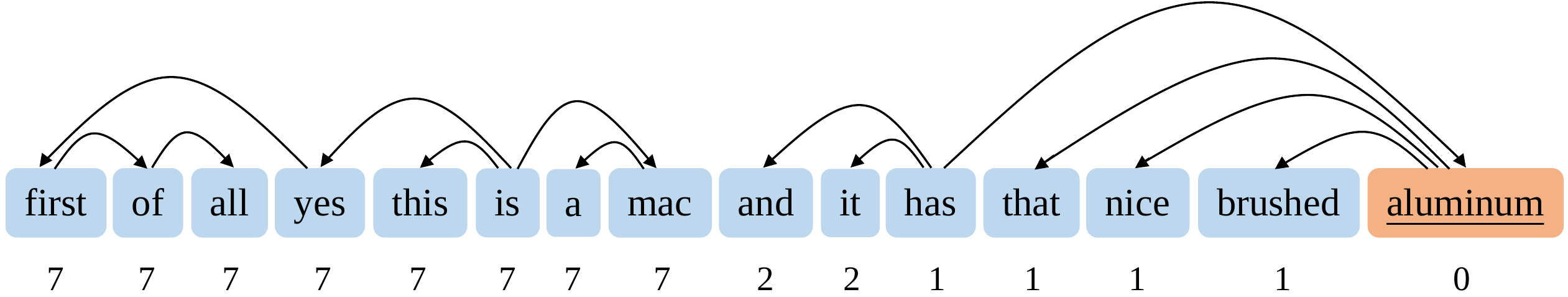}
\caption{Dependency distance with respect to \textit{aluminum}.}
\label{fig:fig2}
\end{figure}

For a better illustration of the proposed method, an example sentence is shown in Figure~\ref{fig:fig2}. With the above described approach, the sequence of tree-based distances for all words in the sentence with respect to the aspect term \textit{aluminum}, $\mathbf{d}=\{d_0,d_1,\dots,d_{\tau},d_{\tau+1},\dots,\\ d_{\tau+m-1},\dots,d_{n-1}\}$ are marked below the words in the figure. The dependency proximity weights of the sentence are then assigned as:
\begin{equation}
p_i=\begin{cases}
1-\frac{d_i}{n}& 0 \leq i<\tau\quad \text{or}\quad \tau+m \leq i < n\\
0& \tau \leq i < \tau+m\\
\end{cases}
\end{equation}

\subsection{Proximity-Weighted Convolution}

Compared with the use of word-level features, Aspect-level sentiment classification with phrase-level features have been shown more effective~\cite{P18-1087, Fan:2018:CMN:3209978.3210115}. We are thus inspired to propose a \textit{proximity-weighted convolution}, which is essentially 1-dimensional convolution with a length-$l$ kernel, i.e. $l$-gram. Different from the original definition of convolution, the proximity-weighted convolution assigns proximity weight before convolution calculation. The proximity weight assigning process is formulated below:
\begin{equation}
r_i=p_i h_i     
\end{equation}

\noindent where $r_i \in \mathbb{R}^{2d_h}$ represents the proximity-weighted representation of the $i$-th word in the sentence.

Additionally, we zero-pad the sentence to ensure the convolution outputs a sequence of the same length as the input sentence. The convolution process contains:
\begin{equation}
t=\left \lfloor \frac{l}{2} \right \rfloor
\end{equation}
\begin{equation}
q_i=\max(\mathbf{W}_{c}^{\mathsf{T}}[r_{i-t}\oplus\cdots\oplus r_i\oplus\cdots\oplus r_{i+t}]+b_c,0)
\end{equation}

\noindent where $q_i \in \mathbb{R}^{2d_h}$ denotes the features extracted by convolution layer, and $\mathbf{W}_{c} \in \mathbb{R}^{l\cdot 2d_h\times 2d_h}$ and $b_c \in \mathbb{R}^{2d_h}$ are weight and bias of the convolution kernel, respectively.

As only few output features of the convolution layer are expected to be instructive for classification,  we choose the most prominent feature $q_s\in \mathbb{R}^{2d_h}$ through a 1-dimension max-pooling layer with a kernel of length $n$, such that:
\begin{equation}
q_s=[\max \limits_{0\leq i <n}q_{i,j}]^{\mathsf{T}}\quad 0\leq j < 2d_h
\end{equation}

\noindent where $q_{i,j}$ is the $j$-th element of $q_i$.

Finally, the most prominent feature vector $q_s$ is fed to a fully connected layer, followed by a softmax normalization to obtain the distribution $y\in \mathbb{R}^{d_p}$ over the decision space on $d_p$-way sentiment polarity:
\begin{equation}
y={\rm softmax}(\mathbf{W}_f^{\mathsf{T}} q_s+b_f)
\end{equation}

\noindent where $b_f \in \mathbb{R}^{d_p}$ is the bias of the fully connected layer and $W_f\in\mathbb{R}^{2d_h\times d_p}$ is the learned weight. 

Our model is trained by the standard gradient descent algorithm, with the loss being the cross entropy loss with $L_2$ regularization:
\begin{equation}
\quad L=-\sum_{(\mathbf{S},\hat{y})\in\mathbf{D}}\sum_u \hat{y}_u \log y_u + \lambda \left\|\Theta\right\|_2
\end{equation}

\noindent Here, $\hat{y}$ means one-hot vector of golden label while $\mathbf{D}$ is the collection of (sentence, label) pairs. And $\Theta$ denotes all trainable parameters, $\lambda$ is the coefficient of $L_2$ regularization.
\section{Experiments}

\subsection{Datasets and Experimental Settings}

We conduct experiments on two benchmarking datasets from SemEval 2014~\cite{S14-2004}. The datasets consist of reviews and comments from two categories: laptop and restaurant, respectively. 

In all of our experiments, 300-dimensional GloVe is leveraged to initialize word embedding~\cite{D14-1162}. All parameters of our model are initialized with the uniform distribution. The dimensionality of hidden state vectors is set to 300. We use Adam as the optimizer with a learning rate of 0.001. The coefficient of $L_2$ regularization is $10^{-5}$ and batch size is 64. We adopt Accuracy and Macro-Averaged F1 as the evaluation metrics. Additionally, the length of n-gram is set to 3\footnote{We have tried several numbers and 3 performed the best.}.

\subsection{Model Comparison}

A comprehensive comparison is carried out between our proposed models, i.e. PWCN with position proximity (\textbf{PWCN-Pos}) and with dependency proximity (\textbf{PWCN-Dep}), against several state-of-the-art baseline models, as listed below:

\begin{itemize}
    \item \textbf{LSTM}~\cite{C16-1311} only uses the last hidden state vector to predict sentiment polarity.
    \item \textbf{RAM}~\cite{D16-1021} considers hidden state vectors of context as external memory and applies Gated Recurrent Unit (GRU) structure to multi-hop attention. The top-most representation is used for predicting polarity.
    \item \textbf{IAN}~\cite{Ma:2017:IAN:3171837.3171854} models attention between aspect and its context interactively with two LSTMs.
    \item \textbf{TNet-LF}~\cite{P18-1087} leverages Context-Preserving Transformation to preserve and strengthen the informative part of context. It also benefits from a multi-layer architecture.
\end{itemize}

We also present comparison with two variants of \textbf{PWCN-Pos}. Firstly, we propose \textbf{Att-PWCN-Pos} model, in which the proximity weight is multiplied by the normalized attention weight, to check whether semantic relatedness and syntax relationship could be incorporated with each other. Further, we intend to measure the effectiveness of n-gram via setting $l$-gram to 1-gram, which naturally degrades convolution process to point-wise feed-forward network, and we call it \textbf{Point-PWCN-Pos}.

\begin{table*}[]
\centering
\caption{Experimental results. Average accuracy and macro-F1 score over 3 runs with random initialization. The best results are in bold. The marker $\dagger$ refers to $p$-value < 0.05 when comparing with IAN, while the marker $\ddagger$ refers to $p$-value < 0.05 when comparing with TNet-LF. The relative increase over the LSTM baseline is given in bracket.}
\label{tab:tab1}
\begin{tabular}{lllll}
\toprule
\multicolumn{1}{c}{\multirow{2}{*}{Model}} & \multicolumn{2}{c}{Laptop} & \multicolumn{2}{c}{Restaurant} \\ 
\cmidrule{2-5}
\multicolumn{1}{c}{}                       & Acc       & Macro-F1       & Acc         & Macro-F1         \\
\midrule
LSTM                                       & 69.63         &    63.51            & 77.99            & 66.91\\
RAM                                     & 72.81 (+4.57\%)          & 68.59 (+8.00\%)               & 79.89 (+2.44\%)           & 69.49 (+3.86\%)\\
IAN                                        & 71.63 (+2.87\%)          & 65.94 (+3.83\%)              & 78.59 (+0.77\%)           & 68.41 (+2.24\%)\\
TNet-LF                                    & 75.16 (+7.94\%)          & 71.10 (+11.95\%)              & 80.20 (+2.83\%)                & 70.78 (+5.78\%)\\
\midrule
Att-PWCN-Pos                                & 72.92 (+4.72\%)          & 68.14 (+7.29\%)              & 80.15 (+2.77\%)           & 70.17 (+4.87\%)\\
Point-PWCN-Pos                              & 74.45 (+6.92\%)         & 69.47 (+9.38\%)              & 80.00 (+2.58\%)           & 69.93 (+4.51\%)\\
\midrule
PWCN-Pos                               & 75.23\textsuperscript{$\dagger$} (+8.17\%)         & 70.71\textsuperscript{$\dagger$} (+11.34\%)              & \textbf{81.12}\textsuperscript{$\dagger \ddagger$} (+4.01\%)           & 71.81\textsuperscript{$\dagger$} (+7.32\%)\\
PWCN-Dep                             & \textbf{76.12}\textsuperscript{$\dagger \ddagger$} (+9.32\%)          & \textbf{72.12}\textsuperscript{$\dagger \ddagger$} (+13.56\%)              & 80.96\textsuperscript{$\dagger$} (+3.81\%)          & \textbf{72.21}\textsuperscript{$\dagger$} (+7.92\%)\\
\bottomrule
\end{tabular}
\end{table*}

\subsection{Experimental Results}

The experimental results in Table \ref{tab:tab1} are yielded by averaging the performances of 3 runs with random initialization. The results demonstrate the general effectiveness of PWCN, which largely outperforms LSTM, RAM and IAN, and also achieves some increase over TNet-LF, the best-performing baseline model under comparison. Among the two types of underlying syntactic structure of sentences captured by PWCN model, dependency proximity brings more benefits to the overall performance than position proximity, with consistently higher Macro-F1 scores on both datasets. The results also support our claim that n-gram information is critical for feature extraction, which can be observed from the disparity between Point-PWCN-Pos and PWCN-Pos. 

Moreover, it is interesting to see that PWCN-based methods with solely syntactic information outperform the Att-PWCN model that combines syntactic and semantic information. While this shows the superiority of leveraging syntactical dependency information to using semantic relatedness, we further conjecture that the attention mechanism could erroneously render term dependencies thus adversely affect the correct decisions of PWCN.

\subsection{Impact of Syntax}

To understand the effect proximity weight has brought, we conduct a case study on an example which could be seen in Table~\ref{tab:tab2}. More specifically, we visualize the weights given by attention in Att-PWCN-Pos, position proximity in PWCN-Pos, and dependency proximity in PWCN-Dep separately along with their predictions.

\begin{table}[]
\centering
\caption{Visualization of a case with respect to \textit{food}}
\label{tab:tab2}
\begin{tabular}{lll}
\toprule
Method & Visualization & Pred. \\
\midrule
Att. & {\setlength{\fboxsep}{0pt}\colorbox{white!0}{\parbox{0.3\textwidth}{
\colorbox{orange!9.229999999999999}{\strut great} \colorbox{orange!36.83}{\strut food} \colorbox{orange!55.15}{\strut but} \colorbox{orange!23.02}{\strut the} \colorbox{orange!100.0}{\strut service} \colorbox{orange!85.54}{\strut was} \colorbox{orange!82.72}{\strut dreadful} \colorbox{orange!0.0}{\strut !} 
}}} & negative\\ 
\midrule 
Pos. & {\setlength{\fboxsep}{0pt}\colorbox{white!0}{\parbox{0.3\textwidth}{
\colorbox{orange!100.0}{\strut great} \colorbox{orange!0.0}{\strut food} \colorbox{orange!100.0}{\strut but} \colorbox{orange!83.33333333333334}{\strut the} \colorbox{orange!66.66666666666666}{\strut service} \colorbox{orange!50.0}{\strut was} \colorbox{orange!33.33333333333333}{\strut dreadful} \colorbox{orange!16.666666666666664}{\strut !} 
}}} & positive\\
\midrule
Dep. & {\setlength{\fboxsep}{0pt}\colorbox{white!0}{\parbox{0.3\textwidth}{
\colorbox{orange!100.0}{\strut great} \colorbox{orange!0.0}{\strut food} \colorbox{orange!100.0}{\strut but} \colorbox{orange!66.66666666666666}{\strut the} \colorbox{orange!66.66666666666666}{\strut service} \colorbox{orange!83.33333333333334}{\strut was} \colorbox{orange!66.66666666666666}{\strut dreadful} \colorbox{orange!50.0}{\strut !} 
}}} & positive\\
\bottomrule
\end{tabular}
\end{table}

We can observe that the existing attention mechanism makes wrong decision on which context word depicts \textit{food} in an extreme way, while both sorts of proximity weight in our model handle this problem properly, which is within our expectation. 


%
%

\section{Conclusions and Future Work}

Previous methods of utilizing aspect information for the aspect-level sentiment classification depend on the modelling of aspect representation from a semantic perspective, while the syntactic relationship between the aspect and its context is generally neglected. In this paper, we have built a framework that leverages n-gram information and syntactic dependency between aspect and contextual terms into an applicable model. Experimental results have demonstrated the effectiveness of our proposed models and suggested that syntactic dependency is more beneficial to aspect-level sentiment classification than semantic relatedness.

We believe it is a promising direction to dive into concrete examples to analyze the difference between PWCN models and attention-based models to achieve a deep understanding of where the syntactical dependencies overwhelm semantic relatedness.

\section*{acknowledgements}
This work is supported by The National Key Research and Development Program of China (grant No. 2018YFC0831700), Natural Science Foundation of China (grant No. U1636203), and Major Project Program of Zhejiang Lab (grant No. 2019DH0ZX01).

\bibliographystyle{ACM-Reference-Format}
\bibliography{ref}


\begin{thebibliography}{15}


\ifx \showCODEN    \undefined \def \showCODEN     #1{\unskip}     \fi
\ifx \showDOI      \undefined \def \showDOI       #1{#1}\fi
\ifx \showISBNx    \undefined \def \showISBNx     #1{\unskip}     \fi
\ifx \showISBNxiii \undefined \def \showISBNxiii  #1{\unskip}     \fi
\ifx \showISSN     \undefined \def \showISSN      #1{\unskip}     \fi
\ifx \showLCCN     \undefined \def \showLCCN      #1{\unskip}     \fi
\ifx \shownote     \undefined \def \shownote      #1{#1}          \fi
\ifx \showarticletitle \undefined \def \showarticletitle #1{#1}   \fi
\ifx \showURL      \undefined \def \showURL       {\relax}        \fi
\providecommand\bibfield[2]{#2}
\providecommand\bibinfo[2]{#2}
\providecommand\natexlab[1]{#1}
\providecommand\showeprint[2][]{arXiv:#2}

\bibitem[\protect\citeauthoryear{Bahdanau, Cho, and Bengio}{Bahdanau
  et~al\mbox{.}}{2015}]%
        {BahdanauCB14}
\bibfield{author}{\bibinfo{person}{Dzmitry Bahdanau},
  \bibinfo{person}{Kyunghyun Cho}, {and} \bibinfo{person}{Yoshua Bengio}.}
  \bibinfo{year}{2015}\natexlab{}.
\newblock \showarticletitle{Neural Machine Translation by Jointly Learning to
  Align and Translate}. In \bibinfo{booktitle}{\emph{ICLR}}.
\newblock


\bibitem[\protect\citeauthoryear{Bengio, Ducharme, Vincent, and Jauvin}{Bengio
  et~al\mbox{.}}{2003}]%
        {bengio2003neural}
\bibfield{author}{\bibinfo{person}{Yoshua Bengio}, \bibinfo{person}{R{\'e}jean
  Ducharme}, \bibinfo{person}{Pascal Vincent}, {and} \bibinfo{person}{Christian
  Jauvin}.} \bibinfo{year}{2003}\natexlab{}.
\newblock \showarticletitle{A neural probabilistic language model}.
\newblock \bibinfo{journal}{\emph{Journal of machine learning research}}
  \bibinfo{volume}{3}, \bibinfo{number}{Feb} (\bibinfo{year}{2003}),
  \bibinfo{pages}{1137--1155}.
\newblock


\bibitem[\protect\citeauthoryear{Chen, Sun, Bing, and Yang}{Chen
  et~al\mbox{.}}{2017}]%
        {D17-1047}
\bibfield{author}{\bibinfo{person}{Peng Chen}, \bibinfo{person}{Zhongqian Sun},
  \bibinfo{person}{Lidong Bing}, {and} \bibinfo{person}{Wei Yang}.}
  \bibinfo{year}{2017}\natexlab{}.
\newblock \showarticletitle{Recurrent Attention Network on Memory for Aspect
  Sentiment Analysis}. In \bibinfo{booktitle}{\emph{EMNLP}}.
  \bibinfo{pages}{452--461}.
\newblock


\bibitem[\protect\citeauthoryear{Dong, Wei, Tan, Tang, Zhou, and Xu}{Dong
  et~al\mbox{.}}{2014}]%
        {P14-2009}
\bibfield{author}{\bibinfo{person}{Li Dong}, \bibinfo{person}{Furu Wei},
  \bibinfo{person}{Chuanqi Tan}, \bibinfo{person}{Duyu Tang},
  \bibinfo{person}{Ming Zhou}, {and} \bibinfo{person}{Ke Xu}.}
  \bibinfo{year}{2014}\natexlab{}.
\newblock \showarticletitle{Adaptive Recursive Neural Network for
  Target-dependent Twitter Sentiment Classification}. In
  \bibinfo{booktitle}{\emph{ACL}}. \bibinfo{pages}{49--54}.
\newblock


\bibitem[\protect\citeauthoryear{Fan, Gao, Du, Gui, Xu, and Wong}{Fan
  et~al\mbox{.}}{2018}]%
        {Fan:2018:CMN:3209978.3210115}
\bibfield{author}{\bibinfo{person}{Chuang Fan}, \bibinfo{person}{Qinghong Gao},
  \bibinfo{person}{Jiachen Du}, \bibinfo{person}{Lin Gui},
  \bibinfo{person}{Ruifeng Xu}, {and} \bibinfo{person}{Kam-Fai Wong}.}
  \bibinfo{year}{2018}\natexlab{}.
\newblock \showarticletitle{Convolution-based Memory Network for Aspect-based
  Sentiment Analysis}. In \bibinfo{booktitle}{\emph{SIGIR}}.
  \bibinfo{address}{New York, NY, USA}, \bibinfo{pages}{1161--1164}.
\newblock


\bibitem[\protect\citeauthoryear{He, Lee, Ng, and Dahlmeier}{He
  et~al\mbox{.}}{2018}]%
        {C18-1096}
\bibfield{author}{\bibinfo{person}{Ruidan He}, \bibinfo{person}{Wee~Sun Lee},
  \bibinfo{person}{Hwee~Tou Ng}, {and} \bibinfo{person}{Daniel Dahlmeier}.}
  \bibinfo{year}{2018}\natexlab{}.
\newblock \showarticletitle{Effective Attention Modeling for Aspect-Level
  Sentiment Classification}. In \bibinfo{booktitle}{\emph{COLING}}.
  \bibinfo{pages}{1121--1131}.
\newblock


\bibitem[\protect\citeauthoryear{Jiang, Yu, Zhou, Liu, and Zhao}{Jiang
  et~al\mbox{.}}{2011}]%
        {Jiang:2011:TTS:2002472.2002492}
\bibfield{author}{\bibinfo{person}{Long Jiang}, \bibinfo{person}{Mo Yu},
  \bibinfo{person}{Ming Zhou}, \bibinfo{person}{Xiaohua Liu}, {and}
  \bibinfo{person}{Tiejun Zhao}.} \bibinfo{year}{2011}\natexlab{}.
\newblock \showarticletitle{Target-dependent Twitter Sentiment Classification}.
  In \bibinfo{booktitle}{\emph{ACL}}. \bibinfo{pages}{151--160}.
\newblock


\bibitem[\protect\citeauthoryear{Li, Bing, Lam, and Shi}{Li
  et~al\mbox{.}}{2018}]%
        {P18-1087}
\bibfield{author}{\bibinfo{person}{Xin Li}, \bibinfo{person}{Lidong Bing},
  \bibinfo{person}{Wai Lam}, {and} \bibinfo{person}{Bei Shi}.}
  \bibinfo{year}{2018}\natexlab{}.
\newblock \showarticletitle{Transformation Networks for Target-Oriented
  Sentiment Classification}. In \bibinfo{booktitle}{\emph{ACL}}.
  \bibinfo{pages}{946--956}.
\newblock


\bibitem[\protect\citeauthoryear{Luong, Pham, and Manning}{Luong
  et~al\mbox{.}}{2015}]%
        {D15-1166}
\bibfield{author}{\bibinfo{person}{Thang Luong}, \bibinfo{person}{Hieu Pham},
  {and} \bibinfo{person}{Christopher~D. Manning}.}
  \bibinfo{year}{2015}\natexlab{}.
\newblock \showarticletitle{Effective Approaches to Attention-based Neural
  Machine Translation}. In \bibinfo{booktitle}{\emph{EMNLP}}.
  \bibinfo{pages}{1412--1421}.
\newblock


\bibitem[\protect\citeauthoryear{Ma, Li, Zhang, and Wang}{Ma
  et~al\mbox{.}}{2017}]%
        {Ma:2017:IAN:3171837.3171854}
\bibfield{author}{\bibinfo{person}{Dehong Ma}, \bibinfo{person}{Sujian Li},
  \bibinfo{person}{Xiaodong Zhang}, {and} \bibinfo{person}{Houfeng Wang}.}
  \bibinfo{year}{2017}\natexlab{}.
\newblock \showarticletitle{Interactive Attention Networks for Aspect-level
  Sentiment Classification}. In \bibinfo{booktitle}{\emph{IJCAI}}.
  \bibinfo{pages}{4068--4074}.
\newblock


\bibitem[\protect\citeauthoryear{Pennington, Socher, and Manning}{Pennington
  et~al\mbox{.}}{2014}]%
        {D14-1162}
\bibfield{author}{\bibinfo{person}{Jeffrey Pennington},
  \bibinfo{person}{Richard Socher}, {and} \bibinfo{person}{Christopher
  Manning}.} \bibinfo{year}{2014}\natexlab{}.
\newblock \showarticletitle{Glove: Global Vectors for Word Representation}. In
  \bibinfo{booktitle}{\emph{EMNLP}}. \bibinfo{pages}{1532--1543}.
\newblock


\bibitem[\protect\citeauthoryear{Pontiki, Galanis, Pavlopoulos, Papageorgiou,
  Androutsopoulos, and Manandhar}{Pontiki et~al\mbox{.}}{2014}]%
        {S14-2004}
\bibfield{author}{\bibinfo{person}{Maria Pontiki}, \bibinfo{person}{Dimitris
  Galanis}, \bibinfo{person}{John Pavlopoulos}, \bibinfo{person}{Harris
  Papageorgiou}, \bibinfo{person}{Ion Androutsopoulos}, {and}
  \bibinfo{person}{Suresh Manandhar}.} \bibinfo{year}{2014}\natexlab{}.
\newblock \showarticletitle{SemEval-2014 Task 4: Aspect Based Sentiment
  Analysis}. In \bibinfo{booktitle}{\emph{SemEval}}. \bibinfo{pages}{27--35}.
\newblock


\bibitem[\protect\citeauthoryear{Tang, Qin, Feng, and Liu}{Tang
  et~al\mbox{.}}{2016b}]%
        {C16-1311}
\bibfield{author}{\bibinfo{person}{Duyu Tang}, \bibinfo{person}{Bing Qin},
  \bibinfo{person}{Xiaocheng Feng}, {and} \bibinfo{person}{Ting Liu}.}
  \bibinfo{year}{2016}\natexlab{b}.
\newblock \showarticletitle{Effective LSTMs for Target-Dependent Sentiment
  Classification}. In \bibinfo{booktitle}{\emph{COLING}}.
  \bibinfo{pages}{3298--3307}.
\newblock


\bibitem[\protect\citeauthoryear{Tang, Qin, and Liu}{Tang
  et~al\mbox{.}}{2016a}]%
        {D16-1021}
\bibfield{author}{\bibinfo{person}{Duyu Tang}, \bibinfo{person}{Bing Qin},
  {and} \bibinfo{person}{Ting Liu}.} \bibinfo{year}{2016}\natexlab{a}.
\newblock \showarticletitle{Aspect Level Sentiment Classification with Deep
  Memory Network}. In \bibinfo{booktitle}{\emph{EMNLP}}.
  \bibinfo{pages}{214--224}.
\newblock


\bibitem[\protect\citeauthoryear{Vo and Zhang}{Vo and Zhang}{2015}]%
        {Vo:2015:TTS:2832415.2832437}
\bibfield{author}{\bibinfo{person}{Duy-Tin Vo} {and} \bibinfo{person}{Yue
  Zhang}.} \bibinfo{year}{2015}\natexlab{}.
\newblock \showarticletitle{Target-dependent Twitter Sentiment Classification
  with Rich Automatic Features}. In \bibinfo{booktitle}{\emph{IJCAI}}.
  \bibinfo{pages}{1347--1353}.
\newblock


\end{thebibliography}

\end{document}